\documentclass[10pt,twocolumn,letterpaper]{article}

\usepackage{cvpr}
\usepackage{times}
\usepackage{epsfig}
\usepackage{graphicx}
\usepackage{amsmath}
\usepackage{amssymb}
\usepackage{amsfonts}
\usepackage{bm}

\usepackage[breaklinks=true,bookmarks=false]{hyperref}

\cvprfinalcopy 

\def\cvprPaperID{1517} 

\ifcvprfinal\fi
\begin{document}

\title{Lie Algebrized Gaussians for Image Representation}

\author{Liyu Gong, Meng Chen and Chunlong Hu\\
  School of CS, Huazhong University of Science and
  Technology\\
  {\tt \footnotesize
    \{gongliyu,chenmenghust,huchunlong.hust\}@gmail.com} }
\maketitle

\begin{abstract}
    We present an image representation method which is derived from
    analyzing Gaussian probability density function (\emph{pdf}) space
    using Lie group theory. In our proposed method, images are modeled
    by Gaussian mixture models (GMMs) which are adapted from a
    globally trained GMM called universal background model (UBM).
    Then we vectorize the GMMs based on two facts: (1) components of
    image-specific GMMs are closely grouped together around their
    corresponding component of the UBM due to the characteristic of
    the UBM adaption procedure; (2) Gaussian \emph{pdf}s form a Lie
    group, which is a differentiable manifold rather than a vector
    space. We map each Gaussian component to the tangent vector space
    (named Lie algebra) of Lie group at the manifold position of
    UBM. The final feature vector, named Lie algebrized Gaussians
    (LAG) is then constructed by combining the Lie algebrized Gaussian
    components with mixture weights. We apply LAG features to scene
    category recognition problem and observe state-of-the-art
    performance on 15Scenes benchmark.
\end{abstract}

\section{Introduction}
Image representation (feature) is one of the most important tasks in
computer vision. Recently Gaussian mixture models (GMMs), which have
been widely used for audio representation
\cite{Reynolds-DSP-2000-UBM-MAP} in speech recognition community, have
been adopted to describe images
\cite{Yan-CVPR-2008-GMM-AGE}\cite{Zhou-ICCV09-HG}. Compared with the
popular histogram image representation, GMMs have some attractive
advantages ( \eg soft assignment, flexible to capture spatial
information) and show better performance in many visual recognition
applications
\cite{Perronnin-ECCV-2006-GMM-Video}\cite{Yan-CVPR-2008-GMM-AGE}\cite{Zhou-ICCV09-HG}\cite{Wu-CVPR-2011-GMM-ActionRecognition}. One
of the major problems of GMMs is that they do not form a vector space
and can not convert to vectors trivially. Various vectorization
methods for GMM representation have been developed in speech
recognition community
\cite{Reynolds-DSP-2000-UBM-MAP}\cite{Campbell-SPL-2006-GMMSupvec}\cite{Lee-SPL-2009-GMMSupvec-Bhattach}
and adopted to image classification applications
\cite{Zhou-ICCV09-HG}. The problem is clear: mapping elements in a
space formed by Gaussian probability density functions (\emph{pdf}s)
to a vector space. However, none of the existing solutions take the
properties of Gaussian function space into consideration. To do so, a
fundamental question should be answered: what kind of space do
Gaussian \emph{pdf}s form? Recently, Gong \etal
\cite{Gong-CVPR-2009-GaussianLieGroup} theoretically point out that
Gaussian \emph{pdf}s are isomorphic to a special kind of affine
matrices which form a Lie group. A Lie group is a differentiable
manifold which is different from ordinary vector spaces. The structure
of the manifold can be analyzed using Lie group theory. Therefore, we
can vectorize GMMs to more effective image descriptors by taking the
Lie group properties of Gaussian \emph{pdf} space into consideration.

\begin{figure}[tbp]
    \centering
    \includegraphics[width=.72\linewidth]{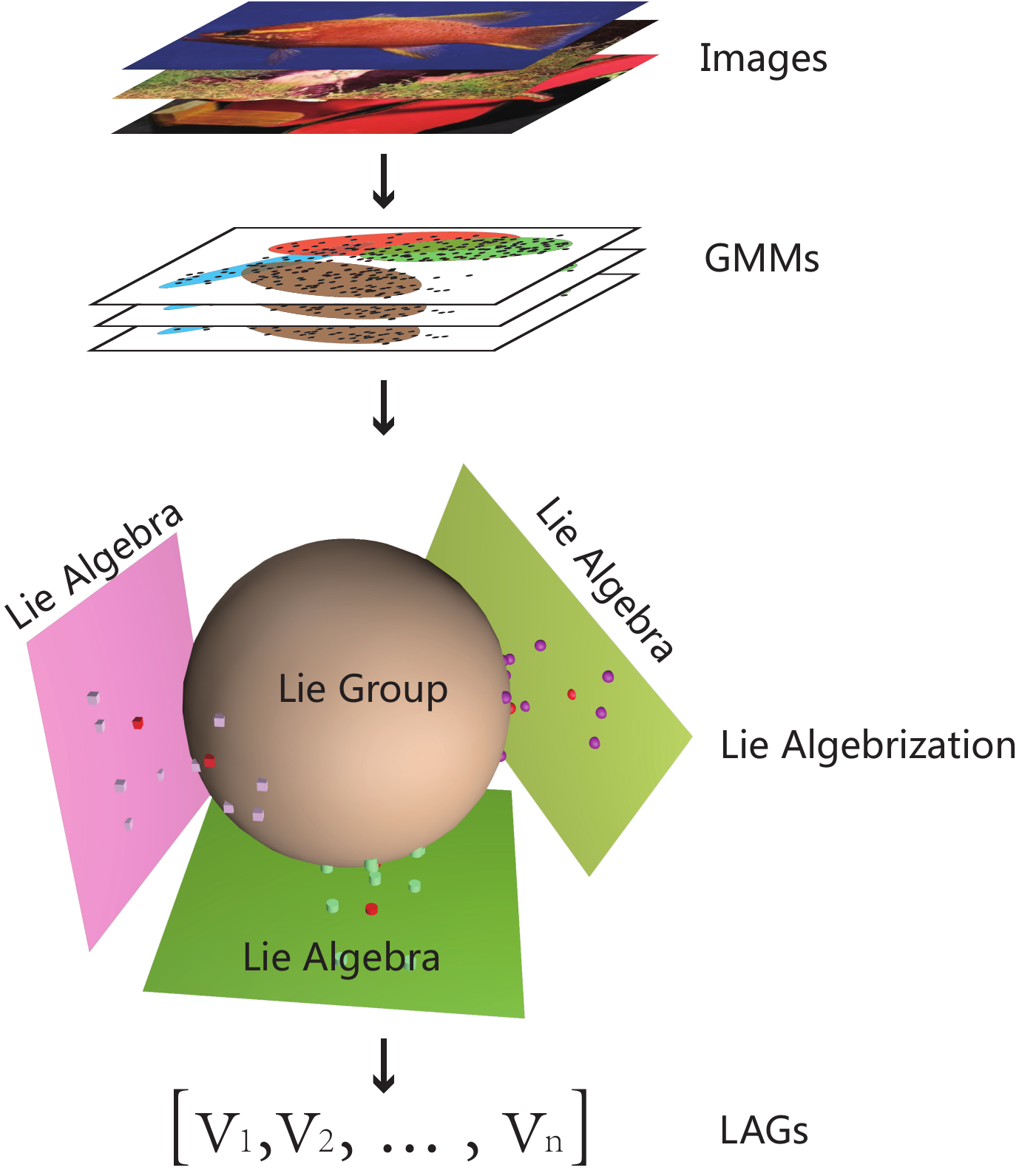}
    \caption{Illustration of LAG feature extraction
      procedure. Firstly, images are modeled by GMMs over local
      patch-level features. Each component of GMMs is represented as a
      point in the Lie group manifold formed by Gaussian
      \emph{pdf}s. Since components from different GMMs are closely
      grouped together, we vectorize them by mapping them to tangent
      space of Lie group. Finally, we combine vectors of each
      component into our final LAG feature.}
    \label{fig:lag}
\end{figure}

In this paper, we propose a novel image representation by
investigating the problem of vectorization GMMs via analyzing Gaussian
\emph{pdf} space. Figure \ref{fig:lag} gives an overview of our
proposed method. The procedure of feature extraction is summarized as
the following four major steps.

\begin{itemize}
\item \emph{First}, images are modeled as GMMs over dense sampled
    patches. We employ a maximum a posteriori (MAP) which is used in
    \cite{Reynolds-DSP-2000-UBM-MAP}\cite{Yan-CVPR-2008-GMM-AGE}\cite{Zhou-ICCV09-HG}
    to estimate the GMMs: We train global GMM called universal
    background model (UBM) on the whole image corpus then adapt it to
    each image. Such a UBM adaptation GMM training approach is much
    more efficient and effective than the ordinary expectation
    maximization (EM) algorithm.
\item \emph{Then}, we parameterize each component of the GMMs to a
    upper triangular definite affine transformation (UTDAT)
    matrix. UTDAT matrices are isomorphic to Gaussian
    \emph{pdf}s. Since UTDAT matrices form a Lie group (which means
    Gaussian \emph{pdf}s form a Lie group too), the Gaussian
    components are points in the Lie group manifold. In figure
    \ref{fig:lag}, the Lie group manifold is represented by a sphere
    surface and Gaussian components are represented by points on the
    surface. Because GMMs trained by UBM-MAP have the same number of
    components as the UBM and each component is just a little shift
    from its corresponding component of the UBM, components are
    closely grouped together around the corresponding component of
    the UBM (represented by red points) in the manifold.
\item \emph{Next}, we utilize characteristic of UBM adapted GMMs to
    vectorize their components (\ie UTDAT matrices) by local
    mapping. To be precise, we map Gaussian components to the tangent
    space of the Lie group manifold at the point of corresponding
    component of the UBM. Since the Gaussian components are locally
    grouped together around the UBM, the mapping preserves the local
    structure of the manifold. The tangent spaces, which are termed as
    \emph{Lie algebra}s, are ordinary vector spaces.
\item \emph{Finally}, we derive our combined vector formula of GMM by
    approximating the inner product of GMMs using a sum of product
    kernel of mixture weights and Lie algebrized components. The final
    vectorized GMM, which is termed as Lie algebrized Gaussians (LAGs),
    is an effective vector representation of the original image and is
    suitable for well known machine learning algorithms.
\end{itemize}

We apply our proposed LAG feature to scene category recognition.
Discriminant nuisance attribute projection (NAP)
\cite{Vogt-Odyssey-2008-Discriminant-NAP} is employed to reduce the
intra-class variabilities. Then a simple nearest centroid (NC)
classifier is adopted to perform the classification task. Our method
show better performance than state-of-the-art methods on 15Scenes
benchmark dataset. To be precise, we get $88.4\%$ average
accuracy. Furthermore, experimental results show that our Lie
algebrization approach is superior to the widely used KullbackLeibler
(KL) divergence based vectorization method.

The remaining of this paper is arranged as follows. In section
\ref{sec:related-work}, we present a short review of the related
work. Section \ref{sec:lie-algebr-gauss} describe the technical detail
of LAG. In section \ref{sec:image-class-using}, the method for image
classification using LAG feature is given. Experimental results on
15Scenes dataset are reported in section
\ref{sec:experimental-results}. Conclusions are made and some future
research issues are given.

\section{Related Work}
\label{sec:related-work}

In recent years bag-of-features (BoF) image representation has been
widely investigated in visual recognition systems. Inspired by the
bag-of-word (BoW) idea in text information retrieval, BoF treats an
image as an collection of local feature descriptors extracted at
densely sampled patches or sparse interest points, encodes them into
discrete ``visual words" using K-means vector quantization (VQ), then
builds a histogram representation of these visual words
\cite{Sivic-PAMI-2009-BoF}. One major problem of BoF approach is that
spatial order of the local descriptors is discarded. Spatial
information is important for many visual recognition applications (\eg
scene categorization and object recognition). To overcome this
problem, Lazebnik \etal propose a BoF extension called spatial pyramid
matching (SPM) \cite{Lazebnik-CVPR-2006-SPM}. In the SPM approach, an
image is partitioned into $2^l\times{}2^l$ sub-images at different
scale level $l=0,1,2\ldots$. Then BoF histogram is computed for each
sub-image. Finally, a vector representation is formed by concatenating
all the BoF histograms. Because of its remarkable improvements on
several image classification benchmarks like Caltech-101
\cite{Fei-Fei-2004-Caltech-101} and Caltech-256
\cite{Griffin-2007-Caltech-256}, SPM has become a standard component
in most image classification systems.

On the other hand, GMMs are widely used for speech signal
representation and have become a standard component in most speaker
recognition systems
\cite{Lee-SPL-2009-GMMSupvec-Bhattach}\cite{Campbell-SPL-2006-GMMSupvec}\cite{Reynolds-DSP-2000-UBM-MAP}. In
GMM based speech signal representation, low-level features are
extracted at local audio segments, then a GMM is estimated on these
features for each speech clip. Reynolds \etal
\cite{Reynolds-DSP-2000-UBM-MAP} propose a novel GMM training method
called universal background model (UBM) adaptation. UBM adaptation
employ a maximum a posterior (MAP) approach instead of normally used
maximum likelihood (ML) approach, \eg expectation maximization
(EM). UBM adaption produces more discriminative GMM representations
and is more efficient than ML estimation. Compared with BoF histogram
representation, GMMs encode the local features in a continuous
probability distribution using soft assignment instead of hard vector
quantization. Zhou \etal \cite{Zhou-ICCV09-HG} adopt GMM to image
representation and report superior performance than SPM in several
image classification applications. The problem of GMM representation
is that GMMs do not form a vector space and can not be converted to
vectors trivially. To get effective vector representation, one should
vectorize GMMs according to the structure properties of the space they
formed. However, none of the existing approaches (\eg
\cite{Lee-SPL-2009-GMMSupvec-Bhattach}\cite{Campbell-SPL-2006-GMMSupvec})
take the structure properties of GMM space into consideration.

Feature space structure analysis is a new computer vision topic
investigated in recent years. Tuzel \etal
\cite{Tuzel-PAMI-2008-CovCascade} analyze the space structure formed
by covariance matrices in a cascade based object detection scenario. A
boosting algorithm is used to train the node classifiers on covariance
features for the detection cascade. Since covariance matrices form a
Riemannian manifold, they are mapped to tangent space at their mean
point before feeding to the weak learner of each boosting
iteration. Compared with treating covariance matrices as vectors
trivially, significant improvement is gained by taking the Riemannian
manifold property of covariance feature space into consideration
during machine learning. Gong \etal
\cite{Gong-CVPR-2009-GaussianLieGroup} derive a Lie group distance
measure for Gaussian \emph{pdf}s by analyzing the structure of a
special kind of affine transformation matrix which is isomorphic to
Gaussian \emph{pdf}. It has been found empirically that Lie group
based Gaussian \emph{pdf} distance is superior to the widely used
Kullback-Leibler (KL) divergence
\cite{Gong-ICIMCS-2009-GMM-LGEMD}\cite{Gong-ICME-2009-Spatiogram-LieGroup}.

In this paper, we derive a feature descriptor by analyzing UBM adapted
GMMs using Lie group theory. Compared with covariance
\cite{Tuzel-PAMI-2008-CovCascade} and Gaussian descriptor
\cite{Gong-CVPR-2009-GaussianLieGroup}, our proposed LAG descriptor is
a kind of holistic descriptors rather than local descriptors. Compared
with previous GMM based audio and image representation method, our
proposed method takes the structural properties of UBM adapted GMMs
into consideration. Experiment results on scene recognition prove the
effectiveness of our method.

\section{Lie Algebrized Gaussians for Image Representation}
\label{sec:lie-algebr-gauss}
\subsection{Image Modeling Using UBM adapted GMM}
\label{sec:image-modeling-using}
We extract local features within densely sampled patches and represent
an image using the probability distribution of its local
features. Specifically, kernel descriptors \cite{Bo-NIPS-2009-KDES}
are computed for each patch. The distribution of local features within
an image are modeled by a GMM. Let $s$ denote patch-level feature
vector. The \emph{pdf} of $s$ is modeled as
\begin{equation}
    \label{eq:1}
    p(s|\Theta) = \sum_{k=1}^{K}\omega_k\bm{\mathcal{N}}(s;\mu_k,\Sigma_k)
\end{equation}
where $K$ denotes number of Gaussian components. $\mathcal{N}$ is
multivariate normal \emph{pdf}. $\omega_k$, $\mu_k$ and $\Sigma_k$ are
the weight, mean vector and covariance matrix of the $k_{th}$
component. For efficiency consideration, we restrict $\Sigma_k$ to be a
diagonal
matrix. $\Theta\equiv\{\omega_k,\mu_k,\Sigma_k\}_{k=1,2,\ldots{},K}$
denotes the whole parameter set of GMM.

The descriptive capability of GMM increases with the number of
Gaussian components $K$. Normally, hundreds of Gaussians are required
to build an effective representation. Compared with the number of
parameters, however, the number of patches is small and insufficient
to train a GMM using a conventional EM approach. Moreover, EM is
time-consuming for GMMs with hundreds of Gaussians. To overcome these
problems, we employ a UBM adaptation approach
\cite{Reynolds-DSP-2000-UBM-MAP} to estimate the parameters. The
adaptation contains two steps: Firstly, a global GMM (\ie UBM) is
trained using patches from the training set. Then, the parameters of
each image-specific GMM are adapted from the UBM using a one iteration
maximum a posterior approach as follows
\begin{align}
    \label{eq:2}
    \omega_k &= [\alpha_kn_k/T+(1-\alpha_k)\bm{\bar{\omega}}_k]\gamma\\
    \mu_k &= \alpha_k\bm{\mathrm{E}}_k(s)+(1-\alpha_k)\bm{\bar{\mu}}_k\\
    \sigma_k^2 &= \alpha_k\bm{\mathrm{E}}_k(s^2)+(1-\alpha_k)(\bm{\bar{\sigma}}_k^2+\bm{\bar{\mu}}_k^2)-\mu_k^2
\end{align}
where $T$ is the number of patches in a specified
image. $\bm{\bar{\omega}}_k$, $\bm{\bar{\mu}}_k$ and
$\bm{\bar{\sigma}}_k$ are the weight, mean and standard deviation of
the $k_{\mathrm{th}}$ mixture of UBM. $\omega_k$, $\mu_k$ and
$\sigma_k$ are the weight, mean and standard deviation of the
$k_{\mathrm{th}}$ mixture of image-specific GMM. The scale factor,
$\gamma$, is computed over all adapted mixture weights to ensure they
sum to unity. $\alpha_k$ is the adaptation coefficient used to control
the balance between UBM and image-specific GMM.  $n_k$,
$\bm{\mathrm{E}}_k(s)$ and $\bm{\mathrm{E}}_k(s^2)$ are the sufficient
statistics of $s$ used to compute mixture weights, mean and
covariance.
\begin{align}
    \label{eq:3}
    n_k &= \sum_{t=1}^T\bm{\mathrm{Pr}}(k|s_t)\\
    \bm{\mathrm{E}}_k(s) &= \frac{1}{n_k}\sum_{t=1}^T\bm{\mathrm{Pr}}(k|s_t)s_t\\
    \bm{\mathrm{E}}_k(s^2) &= \frac{1}{n_k}\sum_{t=1}^T\bm{\mathrm{Pr}}(k|s_t)s_t^2
\end{align}
where $\bm{\mathrm{Pr}}(k|s_t)$ is the posterior probability that the
$t_{\mathrm{th}}$ patch belongs to the $k_{\mathrm{th}}$ Gaussian
components.
\begin{equation}
    \label{eq:4}
    \bm{\mathrm{Pr}}(k|s_t)=\frac{\bm{\bar{\omega}}\bm{\mathcal{N}}(s_t;
      \bm{\bar{\mu}}_k,\bm{\bar{\sigma}}_k)}
    {\sum_{m=1}^K\bm{\bar{\omega}}_m\bm{\mathcal{N}}(s_t;\bm{\bar{\mu}}_m,\bm{\bar{\sigma}}_m)}
\end{equation}

For each mixture, a data-dependent adaptation coefficient $\alpha_k$
is used, which is defined as
\begin{equation}
    \label{eq:5}
    \alpha_k = \frac{n_k}{n_k+r}
\end{equation}
where $r$ is a fixed control value to give penalty to mixtures with
lower posterior probability.

The parameters of image-specific GMM encode the distributions of
local patch-level features from a specified image, thus can be used as
an effective visual representation of that image. On one hand GMMs are
continuous \emph{pdf}s which can avoid vector quantization problems in
discrete distribution estimation approach such as histograms. But on
the other hand, GMMs are not vectors essentially thus are not suited
to most well-known classifiers, especially linear classifiers. Of
course we may simply concatenate the parameters as vector. But the
structural information of the original GMM space are also regrettably
discarded. The most straightforward way to use the structure
information of GMM feature space is to identify what kind of space it
is and then analyze it using existing theory. Although general GMMs
are complex distributions whose space structure are difficult to be
analyzed, we observe that UBM adapted GMMs have some special
characteristics which can help us to analyze its structure.

\begin{figure}[tbp]
    \centering
    \includegraphics[width=\linewidth]{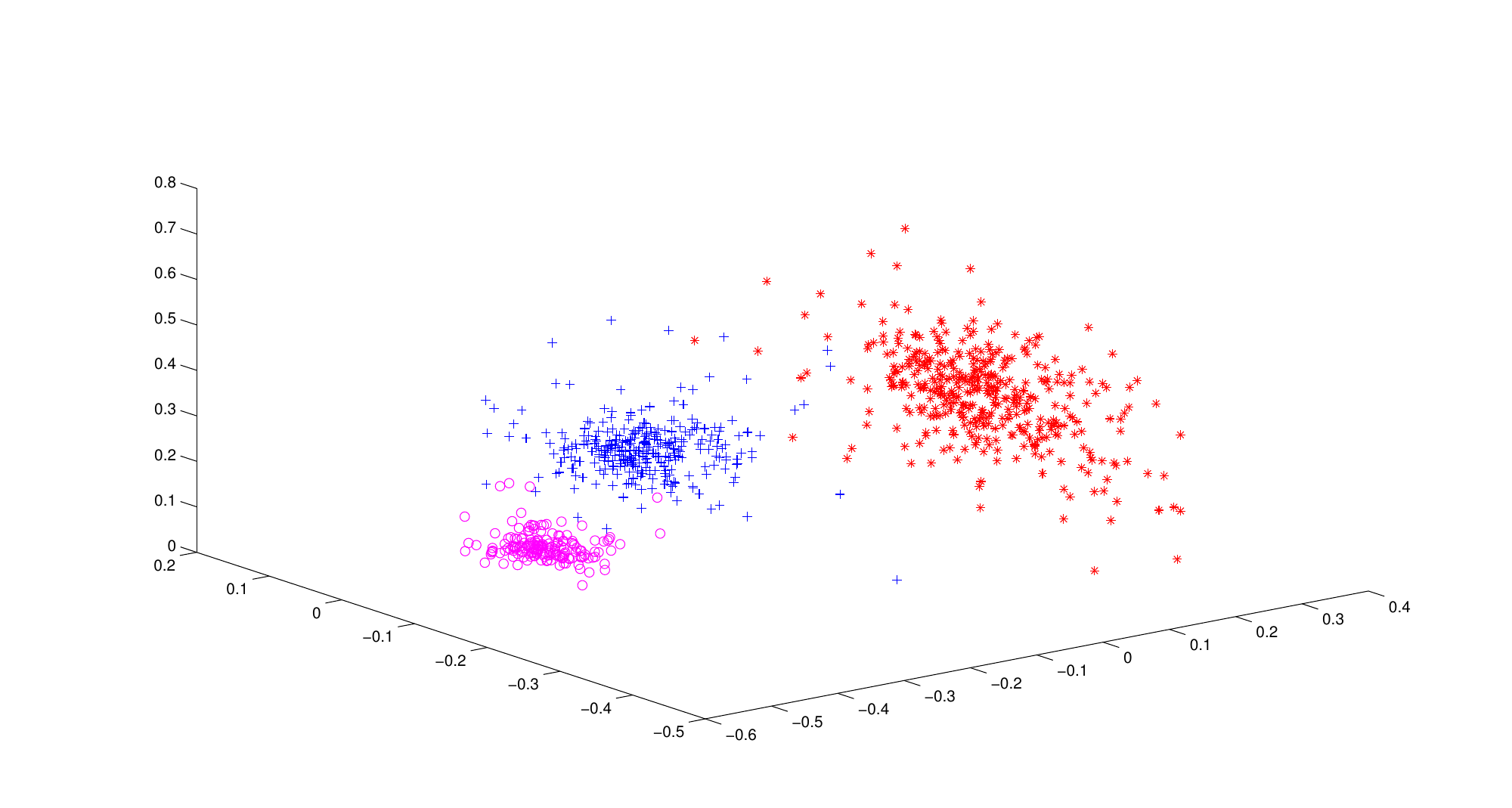}
    \caption{Components of UBM adapted GMMs are closely grouped
      together. We choose $3$ dimension of $3$ components from
      different GMMs and plot them in $3$-d Euclidean space. Each
      point represent a Gaussian component. Different color and marker
    indicate different component index (\ie first, second or third
    component of a GMM).}
    \label{fig:gmm-properties}
\end{figure}

In figure \ref{fig:gmm-properties}, we choose three dimensions of
three components from UBM adapted GMMs trained on 15Scenes dataset and
plot them as points in euclidean space. It can be observed that the
components of these GMMs are closely grouped together around the
components of UBM. Such a characteristic can be explained by the
behavior of UBM adaptation procedure. In UBM adaptation, the MAP
estimation contains one EM-like iteration only. Moreover, a adaptation
coefficient prevents the resultant GMM shift too far from the prior
distribution (\ie UBM) in order to avoid under-fitting. Since
components of image-specific GMMs are closely grouped together, they
have correspondence across images. Therefore, we can analyze Gaussian
components separately then fuse the results together. In the rest of
this section, we show that Gaussian \emph{pdf}s form a Lie group, then
derive a vectorization method for GMM from analyzing Gaussian
\emph{pdf} space using Lie group theory.

\subsection{Gaussian \emph{pdf}s and Lie group}
\label{sec:gauss-emphpdfs-lie}
Let $\bm{\mathrm{x}}_0$ denote a random vector which is standard
multivariate Gaussian distributed (\ie the mean and covariance are
zero vector and identity matrix respectively). Let
$\bm{\mathrm{x}}_1=A\bm{\mathrm{x}}_0+\mu$ be a resultant vector of an
invertible affine transformation from $\bm{\mathrm{x}}_0$. From the
properties of multivariate Gaussian distribution, we can know that
$\bm{\mathrm{x}}_1$ is also multivariate Gaussian
distributed. Furthermore, the mean vector and covariance matrix of
$\bm{\mathrm{x}}_1$ are $\mu$ and $AA^T$ respectively. More generally
speaking, any invertible affine transformation can produce a
multivariate Gaussian distribution. Furthermore, if we restrict $A$ to
be upper triangular and definite, we can get an unique $A$ given a
arbitrary multivariate distribution by Cholesky decomposition, which
means there is a bijection between Gaussian \emph{pdf}s and upper
triangular definite affine transformation (UTDAT). Therefore, Gaussian
\emph{pdf}s are isomorphic to UTDATs. Let $M$ denote the matrix form of
UTDAT which is defined as follow.
\begin{equation}
    \label{eq:6}
    M =
    \begin{bmatrix}
        A & \mu\\
        0 & 1
    \end{bmatrix}
\end{equation}
We can analyze $M$ instead of Gaussian \emph{pdf}s.

Invertible affine transformations form a Lie group and matrix
multiplication is its group operator. UTDAT which is a special case of
invertible affine transformation is closed under matrix multiplication
operation. Therefore, UTDAT is a subgroup of Invertible affine
transformation. Since any subgroup of a Lie group is still a Lie group,
UTDAT is a Lie group. In conclusion, Gaussian \emph{pdf}s form a Lie
group.

In mathematics, a Lie group is a group which is also a differentiable
manifold, with the property that the group operations are compatible
with the smooth structure. An abstract Lie group could have many
isomorphic instances. Each of them is an representation of the
abstract Lie group. In Lie group theory, matrix representation
\cite{Varadarajan-Book-1984} is a useful tool for structure
analysis. In our case, UTDAT is the matrix representation of the
abstract Lie group formed by Gaussian \emph{pdf}s. Specially,
covariance matrices of our GMMs are diagonal thus $A$ is diagonal
too. Precisely, UTDAT of the $k_{\mathrm{th}}$ component is defined as
follow.
\begin{equation}
    \label{eq:6}
    M_k =
    \begin{bmatrix}
      \sigma_{k1} &           &          &        & \mu_{k1}\\
               & \sigma_{k2}  &          &        & \mu_{k2}\\
               &           & \sigma_{k3} &        & \mu_{k3}\\
               &           &          & \ddots & \vdots\\
      0        &         0 &        0 &        & 1
    \end{bmatrix}
\end{equation}
where $\mu_k(d)$ and $\sigma_k(d)$ are the mean and standard deviation
of the $d_{\mathrm{th}}$ dimension of the $k_{\mathrm{th}}$ Gaussian
component respectively.

\subsection{Lie Algebrization of Gaussian components}
\label{sec:lie-algebr-gauss-1}

As discussed before, components of UBM adapted GMM are closely grouped
together around its corresponding component of UBM thus we can
preserve most of their structure information by projecting them to the
tangent space of Lie group at the point of the corresponding
component. In mathematics, the tangent space of a manifold facilitates
the generalization of vectors from affine spaces to general manifolds,
since in the latter case one cannot simply subtract two points to
obtain a vector pointing from one to the other. Analogous to a tangent
plane of a sphere, a tangent space of a Lie group is a vector
space. To best preserve the structure information of a collection of
points in a manifold, the target vector space should be the tangent
space at the mean points of the point set
\cite{Tuzel-PAMI-2008-CovCascade}. In our case, UBM is an
approximation of all the image-specific GMMs thus we use components of
UBM as mean Gaussian \emph{pdf}s.

Let $M_k$ and $\bm{\bar{M}}_k$ denote the $k_{\mathrm{th}}$ component
(UTDAT matrix form) of an image-specific GMM and UBM. Let
$\mathrm{m}_k$ denote the corresponding point in the tangent space
projecting from $M_k$. The projection is accomplished via matrix
logarithm.
\begin{equation}
    \label{eq:7}
    \mathrm{m}_k = \log{}(\bm{\bar{M}}_k^{-1}M_k)
\end{equation}
Note that here $\log$ is matrix logarithm rather than element-wise
logarithm of a matrix. Since tangent space of an Lie group is a vector
space, $\mathrm{m}_k$ is a vector thus we can unfold elements of
$\mathrm{m}_k$ to a vector form.

Although we can project Gaussian components using equation
\eqref{eq:7}, it is not efficient. The $\log$ operation in
\eqref{eq:7} requires Schur decomposition of $\bm{\bar{M}}_k^{-1}M_k$
\cite{Davies-2003-SIAMJMAA-FUNM}, which is
time-consuming. Fortunately, covariance matrices of Gaussian
components are diagonal in our case thus we can develop a efficient
scalar form of $\log$.

Here we derive our scalar form of UTDAT matrix logarithm. For diagonal
Gaussian components, each dimension of transformation is
independent. So we analyze the $1$-d case of UTDAT logarithm
first. Here we let $M$ be a $1$-d UTDAT matrix with the form
\begin{equation}
    \label{eq:8}
    M =
    \begin{bmatrix}
        \sigma & \mu\\
        0 & 1
    \end{bmatrix}
\end{equation}
and let $K=M-I$ where $I$ is a $2$-d identity matrix. Using the series
form of matrix logarithm, we have
\begin{align}
    \label{eq:9}
    m &= \log(M)\\
    &= \log(I+K)\\
    & = \sum_{n=1}^{\infty}(-1)^{n-1}\frac{K^n}{n}\\
    & =
    \begin{bmatrix}
        \sum\limits_{n=1}^{\infty}(-1)^{n-1}\frac{(\sigma-1)^n}{n} &
        \sum\limits_{n=1}^{\infty}(-1)^{n-1}\frac{\mu(\sigma-1)^{n-1}}{n}\\
        0 & 0
    \end{bmatrix}\\
    &=
    \label{eq:9-2}
    \begin{bmatrix}
        \log{}(\sigma) & \frac{\mu\log{}(\sigma)}{\sigma-1}\\
        0 & 0
    \end{bmatrix}
\end{align}
Note that we can always scale the matrix using the following equations
in order to make sure the series convergent
\begin{align}
    \label{eq:10}
    \log{}A &= \log(\lambda(I+B))\\
    &= \log(\lambda{}I)+\log(I+B)\\
    &= (\log\lambda)I+\log(B)
\end{align}
Using the above equations, we have
\begin{align}
    \label{eq:11}
    \log(M_1^{-1}M_2) &= \log(
    \begin{bmatrix}
        \sigma_1^{-1} & -\mu_1\sigma_1^{-1}\\
        0 & 1
    \end{bmatrix}
    \begin{bmatrix}
        \sigma_2 & \mu_2\\
        0 & 1
    \end{bmatrix})\\
    \label{eq:11-2}
    & =\log(
    \begin{bmatrix}
        \sigma_1^{-1}\sigma_2 & (\mu_2-\mu_1)\sigma_1^{-1}\\
        0 & 1
    \end{bmatrix})\\
    \label{eq:11-3}
    & =
    \begin{bmatrix}
        \log(\frac{\sigma_2}{\sigma_1}) &
        (\mu_2-\mu_1)\frac{\log(\sigma_2)-\log(\sigma_1)}{\sigma_2-\sigma_1}\\
        0 & 0
    \end{bmatrix}
\end{align}
Note that we use \eqref{eq:9-2} to derive \eqref{eq:11-3} from
\eqref{eq:11-2}. Finally, we can get our projected Gaussian component
$\mathrm{m}_k$ using the above equations.
\begin{equation}
    \label{eq:12}
    \mathrm{m}_k = \\
    \begin{bmatrix}
        \log\frac{\sigma_{k1}}{\bm{\bar{\sigma}}_{k1}} &           &          &        & \frac{(\mu_{k1}-\bm{\bar{\mu}}_{k1})\log\frac{\sigma_{k1}}{\bm{\bar{\sigma}}_{k1}}}{\sigma_{k1}-\bm{\bar{\sigma}}_{k1}}\\
        & \log\frac{\sigma_{k2}}{\bm{\bar{\sigma}}_{k2}}  &          &        & \frac{(\mu_{k2}-\bm{\bar{\mu}}_{k2})\log\frac{\sigma_{k2}}{\bm{\bar{\sigma}}_{k2}}}{\sigma_{k2}-\bm{\bar{\sigma}}_{k2}}\\
        &           &          & \ddots & \vdots\\
        0        &         0 &        0 &        & 0

    \end{bmatrix}
\end{equation}
If we assume that $\sigma$ always equals $\bm{\bar{\sigma}}$ (\ie
adapt mean only and keep covariance unchanged during UBM adaptation),
$\mathrm{m}_k$ is reduced to $\hat{\mathrm{m}}$
\begin{equation}
    \label{eq:20}
    \hat{\mathrm{m}}_k = \\
    \begin{bmatrix}
        \frac{(\mu_{k1}-\bm{\bar{\mu}}_{k1})}{\bm{\bar{\sigma}}_{k1}},
       &
       \frac{(\mu_{k2}-\bm{\bar{\mu}}_{k2})}{\bm{\bar{\sigma}}_{k2}},
       &\ldots
    \end{bmatrix}
\end{equation}
using the fact
\begin{equation}
    \label{eq:21}
    \lim_{x\rightarrow{}0}\frac{\log(1+x)}{x}=1
\end{equation}
Compared with $\mathrm{m}_k$, $\hat{\mathrm{m}}_k$ represents each
dimension using a scalar (while $\mathrm{m}_k$ uses a $2$-d vector) and
discards the covariance information of GMM.

\subsection{Lie Algebrized Gaussians}
\label{sec:lie-algebr-gauss-2}
After vectorization of Gaussian components, we fuse them together to
get a vectorized GMM. We derive our vectorized GMM from a product
kernel. Let $a$ and $b$ denote two GMMs, we use kernel function
$f(a,b)$ defined as follow
\begin{equation}
    \label{eq:13}
    f(a,b) = \sum_{k=1}^Kf_{\omega}(\omega_k^a,\omega_k^b)f_{\mathrm{m}}(\mathrm{m}_k^a,\mathrm{m}_k^b)
\end{equation}
where $f_a$ and $f_b$ are kernel functions for mixture weights and
vectorized Gaussian \emph{pdf}s. Using linear inner product
$f_{\mathrm{m}}(x,y)=x^Ty$ for vectorized
Gaussian \emph{pdf} and $f_w(x,y)= \sqrt{xy}$ for mixture
weights, we get
\begin{align}
    \label{eq:14}
    f(a,b) &=
    \sum_{k=1}^K\sqrt{\omega_k^a}\sqrt{\omega_k^b}(\mathrm{m}_k^a)^T\mathrm{m}_k^b\\
    \label{eq:14.2}
    &=\sum_{k=1}^K(\sqrt{\omega_k^a}\mathrm{m}_k^a)^T(\sqrt{\omega_k^b}\mathrm{m}_k^b)
\end{align}
Using equation \eqref{eq:14.2}, we designed our final vector $V_{lag}$ as
\begin{equation}
    \label{eq:15}
    V_{lag} = [\sqrt{\omega_1}\mathrm{m}_1,\sqrt{\omega_2}\mathrm{m}_2, \ldots , \sqrt{\omega_K}\mathrm{m}_K]
\end{equation}
The final vector $V_{lag}$, which is named Lie algebrized Gaussians
(LAG), is an effective representation of the original image and is
suitable for most known machine learning techniques. If we replace
$\mathrm{m}_k$ with $\hat{\mathrm{m}}_k$ in \eqref{eq:15}, we can get
a reduced LAG (rLAG) vector which has lower dimensionality but less
discriminative.

\section{Scene Category Recognition Using LAG Feature}
\label{sec:image-class-using}
We apply our LAG feature to scene category recognition. Scene
recognition is a typical and important visual recognition problem in
computer vision. Some digital cameras (\eg Sony W170 and Nikon D3/300)
are also starting to include ``Intelligent Scene Recognition'' modules
to help selecting appropriate aperture, shutter speed, and white
balance.

To address the scene recognition problem, we represent each image
using our proposed LAG vector. Since SPM \cite{Lazebnik-CVPR-2006-SPM}
have been proved empirically to be a useful component for various
visual recognition system, we adopt it to our LAG based
representation. Specifically, we divide image into sub-images in the
same manner as SPM and extract LAG features for each sub-image, then
combine these LAG vectors for image representation. Then we reduce
within-class variability of LAGs using discriminant nuisance attribute
projection \cite{Vogt-Odyssey-2008-Discriminant-NAP}. For efficiency
reasons, we employ a simple nearest centroid (NC) classifier to
classify NAP projected LAG features into different scene categories.

\section{Experimental Results}
\label{sec:experimental-results}
We test our method on the 15Scenes dataset
\cite{Lazebnik-CVPR-2006-SPM}. The scene dataset contains fifteen
scene categories, thirteen of them is provided by Fei-Fei \etal in
\cite{Fei-Fei-CVPR-2005-13Scenes}. Each scene category contains about
$400$ images. The size of each image is about $300\times{}250$
pixels. This dataset is the most comprehensive one for scene category
recognition.

We extract kernel descriptors \cite{Bo-NIPS-2009-KDES} on densely
sampled patches for each image. Specifically, three types of kernel
descriptors are used: color, gradient and LBP kernel
descriptors. Large images are resized to be no larger than
$300\times{}300$. $16\times{}16$ and $24\times{}24$ patches with $4$
pixel step are used. The resultant kernel descriptors are reduced to
$50$-d using principal component analysis (PCA). Each $50$-d vector is
then combined with the normalized x-y spatial coordinates of the
center of its patch window. Therefore, the final patch descriptors are
$52$-d which contains both appearance and spatial information of the
patches. We model each image using GMMs with $512$
components. Specifically, we divide each image into $1\times{}1$ and
$2\times{}2$ pyramid-like sub-images and estimate a GMM for each
sub-image ($5$ GMMs for an image in total). The corresponding $5$ LAG
vectors are concatenated to a single vector. To test scene recognition
performance, we randomly select $100$ images from each category for
training and the rest for testing. The experiments are repeated $10$
times and $88.4\%$ average recognition accuracy is obtained. We
assemble the performance of our algorithm and various state-of-the-art
algorithms in table \ref{tab:result} for comparison.
\begin{table}[htp]
    \centering
    \begin{tabular}[c]{|c|c|}
        \hline
        Algorithm & Average Accuracy(\%)\\
        \hline
        Histogram \cite{Fei-Fei-CVPR-2005-13Scenes} & 65.2\\
        SPM \cite{Lazebnik-CVPR-2006-SPM} & 81.4\\
        HG \cite{Zhou-ICCV09-HG} & 85.3 \\
        KDES+LinSVM \cite{Bo-NIPS-2009-KDES} & 81.9\\
        KDES+LapKSVM \cite{Bo-NIPS-2009-KDES} & 86.7\\
        LAG+NC (this paper) & \textbf{88.4}\\
        \hline
    \end{tabular}
    \caption{Performance of different algorithms on 15Scenes
      dataset. Our LAG+NC method achieves the state-of-the-art performance.}
    \label{tab:result}
\end{table}
The results show that our method outperform all the other
algorithms. Kernel descriptor with Laplacian kernel SVM, which is the
second best, obtains $86.7\%$ average recognition accuracy. Note that
our method uses a simple NC classifier rather than kernel
machines. It is also observed that Laplacian kernel SVM boost the
performance of kernel descriptors a lot from linear SVM ($81.9\%$). So
it is interesting to see what kind of kernel SVM can boost the
performance of our LAG representation. However, our method is more
practical because NC classifier is suitable for large scale dataset.
\begin{figure*}[htp]
    \centering
    \begin{tabular}[c]{ccc}
         \includegraphics[width=.31\linewidth]{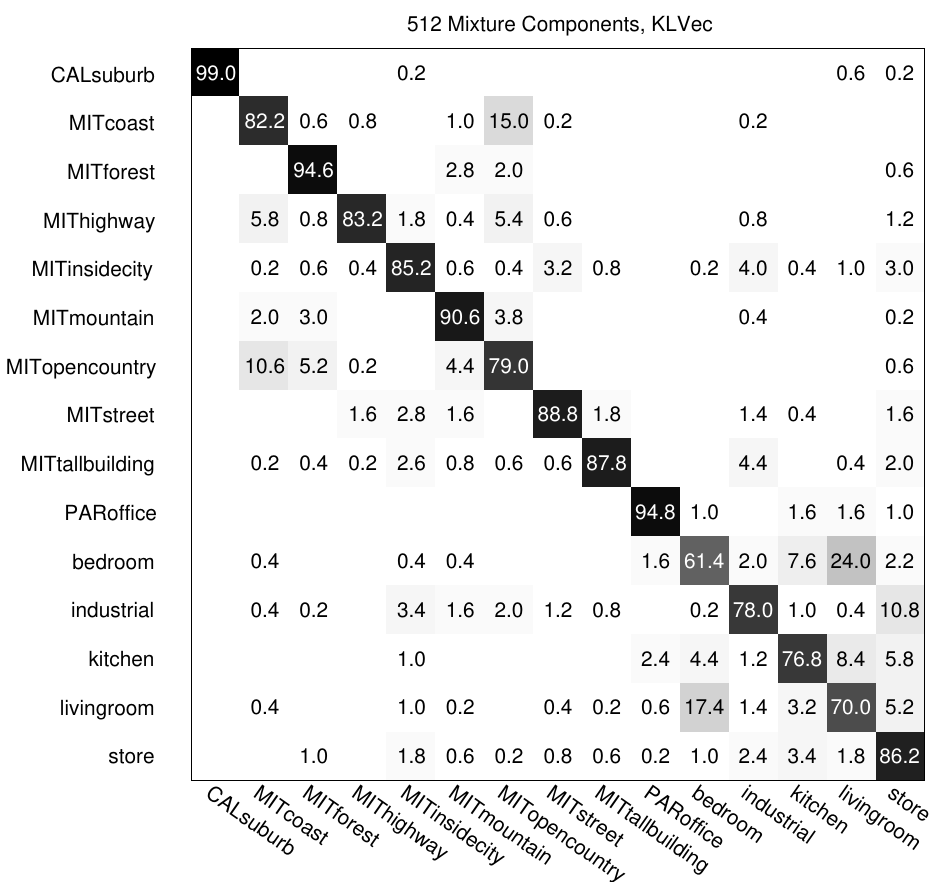}
        &\includegraphics[width=.31\linewidth]{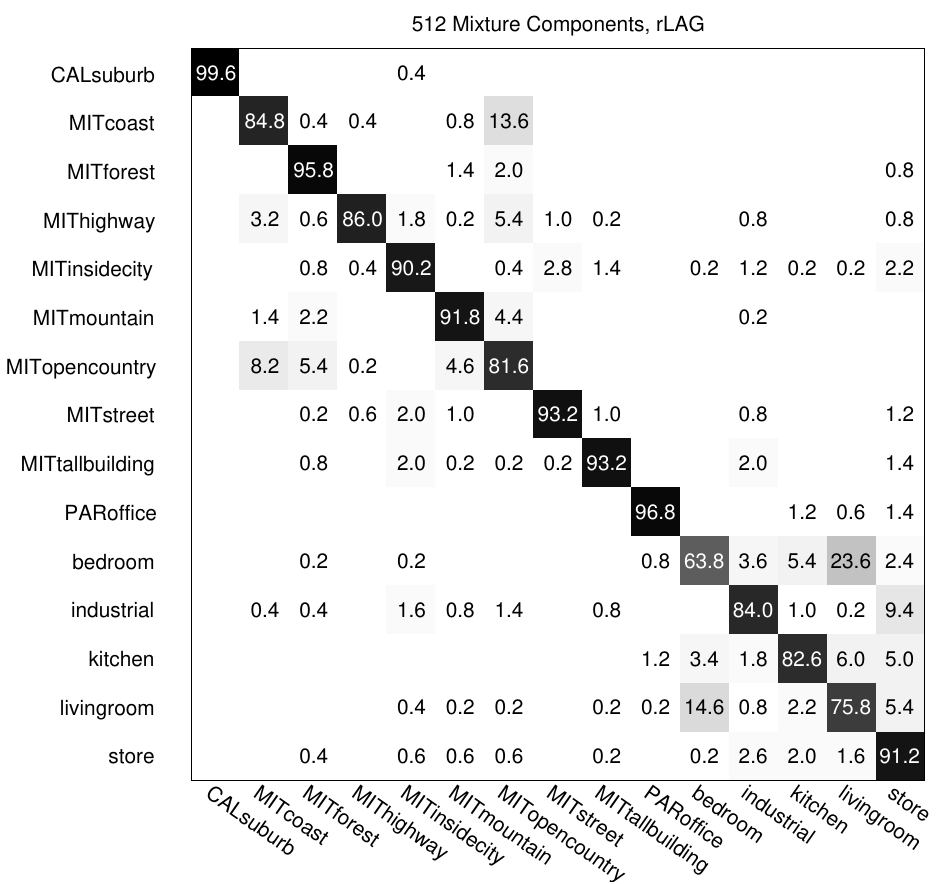}
        &\includegraphics[width=.31\linewidth]{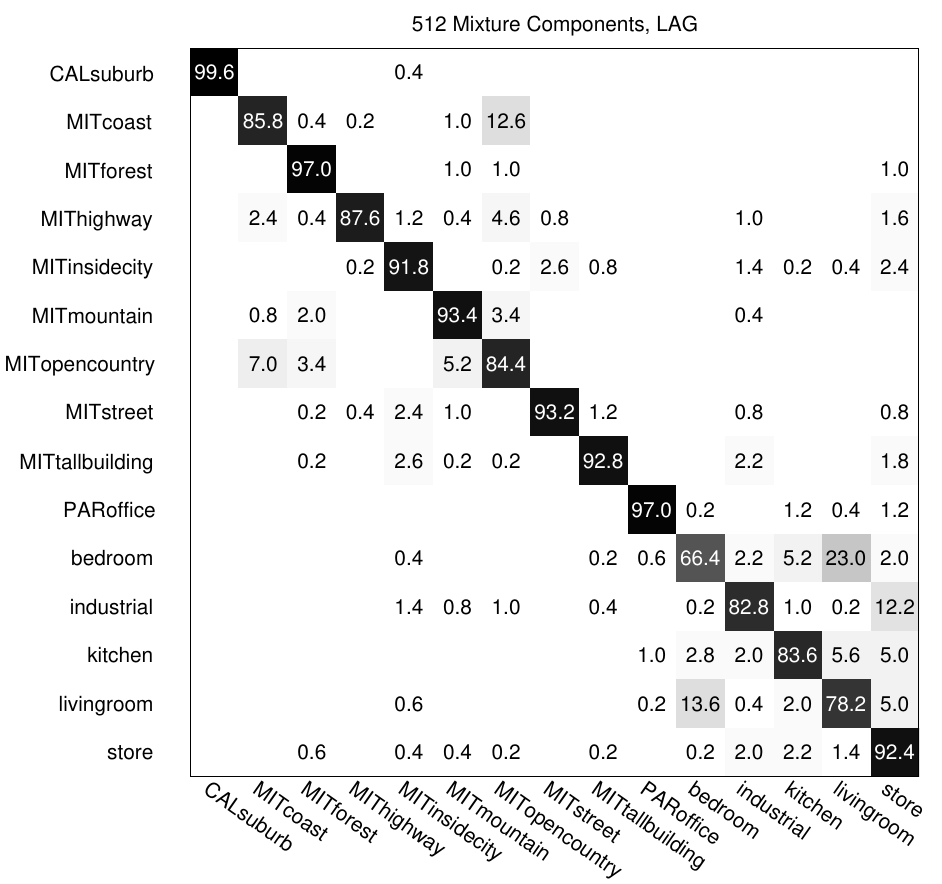}\\
        (a) & (b) & (c)
    \end{tabular}
    \caption{Confusion matrices of the three vectorization approaches:
      (a) KLVec, (b) reduced LAG and (c) LAG. The entry in the
      $i_{\mathrm{th}}$ row and $j_{\mathrm{th}}$ column is the
      percentage of images from the $i_{\mathrm{th}}$ class and
      classified as the $j_{\mathrm{th}}$ class.}
    \label{fig:confmat}
\end{figure*}

The third best algorithm \cite{Zhou-ICCV09-HG} in table
\ref{tab:result} using a KL divergence based vectorization together
with a spatial information scheme called Gaussian map as image
representation. KL divergence based vectorization (KLVec) is a most
widely used GMM vectorization approach and has been empirically proved
to be effective in many applications
\cite{Campbell-SPL-2006-GMMSupvec}\cite{Wu-CVPR-2011-GMM-ActionRecognition}\cite{Yan-CVPR-2008-GMM-AGE}. KLVec
has the following form
\begin{equation}
    \label{eq:16}
    V_{kl}=[\sqrt{\omega_1}\mu_1\bm{\bar{\sigma}}_1^{-1}, \sqrt{\omega_2}\mu_2\bm{\bar{\sigma}}_2^{-1},
    \ldots, \sqrt{\omega_K}\mu_K\bm{\bar{\sigma}}_K^{-1}]
\end{equation}
For a specified dimension (\eg $d_{\mathrm{th}}$) of a specified
component (\eg $k_{\mathrm{th}}$), KLVec encodes the distribution as
\begin{equation}
    \label{eq:18}
    \frac{\mu_{kd}}{\bm{\bar{\sigma}}_{kd}}
\end{equation}
From \eqref{eq:18}, we can clearly see that our LAG feature is
different from KLVec in two aspect: Firstly, KLVec discard covariance
information of GMM. Secondly, mean vector in LAG is centralized by
subtracting the corresponding mean of UBM. Furthermore, the difference
between RLAG and KLVec is mean centralization only.

To compare KLVec with LAG and RLAG empirically, we implement both
KLVec and test it in the same scenario with same parameter settings as
LAG and rLAG. The average accuracies of the three vectors are
presented in table \ref{tab:result2}.

We observe that LAG is significantly superior to KLVec (88.4\% vs
83.8\%). The reduced LAG achives 87.3\% accuracy, which indicates that
the centralization operation of LAG feature is important. The detailed
confusion matrices of the three vectorization approaches are present in
figure \ref{fig:confmat}. Note that our system with KLVec obtain lower
accuracy than the system in \cite{Zhou-ICCV09-HG}, the reason might be
that some components (\eg spatial Gaussian maps) is not included in
our system.

\begin{table}[htp]
    \centering
    \begin{tabular}[c]{|c|c|}
        \hline
        Algorithm & Average Accuracy(\%)\\
        \hline
        KLVec & 83.84 $\pm$ 1.23\\
        rLAG (this paper) & 87.36 $\pm$ 0.95\\
        LAG (this paper)& 88.40 $\pm$ 0.96\\
        \hline
    \end{tabular}
    \caption{Comparison of different vectorization approach. The
      centralization operation of reduced LAG considerably improves the
      performance compared with KLVec. The covariance information in
      LAG vector is also useful for recognition, which improves another
      1\% from reduced LAG feature.}
    \label{tab:result2}
\end{table}

We test the three vectorization approaches (LAG, rLAG, KLVec) with different number of Gaussian mixture components and keep the same setting for the other parameters as described above. The results are shown in table~\ref{tab:gmmk}. According to the table, LAG is always superior to rLAG and KLVec. Moreover, LAG gains fair performance when the number of Gaussian mixture components is just set to 32. This phenomenon demonstrates that the covariance matrix information which is represented by LAG is very useful.

\begin{table*}[t]
\begin{center}
\begin{tabular}{|c|c|c|c|}
\hline
 ~ & \multicolumn{3}{c}{average accuracy(\%)} \vline \\
\hline
Mixture Number & LAG & rLAG & KLVec \\
\hline
32 & 86.41$\pm$0.87\% & 82.59$\pm$1.43\% & 82.59$\pm$1.15\% \\
\hline
64 & 86.88$\pm$1.25\% & 84.76$\pm$1.19\% & 84.27$\pm$1.28\% \\
\hline
128 & 87.55$\pm$1.37\% & 86.11$\pm$1.09\% & 84.57$\pm$1.64\% \\
\hline
256 & 88.17$\pm$1.69\% & 87.03$\pm$1.16\% & 84.24$\pm$1.18\% \\
\hline
512 & 88.40$\pm$0.96\% & 87.36$\pm$0.95\% & 83.84$\pm$1.23\% \\
\hline
1024 & 87.73$\pm$1.38\% & 87.39$\pm$0.86\% & 83.39$\pm$1.30\% \\
\hline
\end{tabular}
\end{center}
\caption{Results for different Gaussian mixture number.}
\label{tab:gmmk}
\end{table*}


\section{Conclusion and Future Work}
\label{sec:concl-future-work}
We analyze the structure of UBM adapted GMMs and derive a Lie group
based GMM vectorization approach for image representation. Since
Gaussian \emph{pdf}s form a Lie group and components of UBM adapted
GMMs are closely grouped together around UBM, we map each component
of a GMM to tangent space (Lie algebra) of Lie group at the position
of corresponding component of UBM. Such a kind of vectorization
approach (named Lie algebrization) preserves the structure of Gaussian
components in the original Lie group manifold. The final Lie
algebrized Gaussians (LAG) features are constructed by combining Lie
algebrized Gaussian components with mixture weights. We apply LAG to
scene category recognition and achieve state-of-the-art performance on
15Scenes benchmark with a simple nearest centroid
classifier. Experimental results also show that our vectorization
approach is considerably superior to the widely used KL divergence
based vectorization method.

There are several interesting issues about LAG based image
representation we shall investigate in the future. Firstly, we shall
apply LAG to other visual recognition problems, such as object
recognition, action recognition. Secondly, it is interesting to
develop a kernel classifier for GMM using its Lie group
structure. Finally, applying LAG feature to audio representation and
comparing it with KL divergence based vectorization is another
interesting topic.

{\small
\bibliographystyle{ieee}
\bibliography{ref}
}

\end{document}